\documentclass{article}

\pdfoutput=1

\usepackage[preprint]{nips_2018}

\usepackage[utf8]{inputenc} 
\usepackage[T1]{fontenc}    
\usepackage{hyperref}       
\usepackage{url}            
\usepackage{booktabs}       
\usepackage{amsfonts}       
\usepackage{nicefrac}       
\usepackage{microtype}      
\usepackage{graphicx}
\usepackage{amsmath}       

\title{Bayesian estimation for large scale multivariate Ornstein-Uhlenbeck model of brain connectivity}

\author{
  Andrea Insabato \\
  Center for Brain and Cognition\\
  University Pompeu Fabra\\
  Barcelona, Spain \\
  \texttt{andrea.insabato@upf.edu} \\
  \And
  John P. Cunningham \\
  Theoretical Neuroscience \\
  Columbia University \\
  New York City, NY, USA \\
  \texttt{jpc2121@columbia.edu} \\
  \And
  Matthieu Gilson \\
  Center for Brain and Cognition\\
  University Pompeu Fabra\\
  Barcelona, Spain \\
  \texttt{matthieu.gilson@upf.edu} \\
}

\begin{document}

\maketitle

\begin{abstract}
	Estimation of reliable whole-brain connectivity is a crucial step towards the use of connectivity
	information in quantitative approaches to the study of neuropsychiatric disorders.
	In estimating brain connectivity a challenge is imposed by the paucity of time samples and the
	large dimensionality of the measurements. Bayesian estimation methods for network models offer a
	number of advantages in this context but are not commonly employed.
	Here we compare three different estimation methods for the multivariate Ornstein-Uhlenbeck model, that
	has recently gained some popularity for characterizing whole-brain connectivity. We first show that
	a Bayesian estimation of model parameters assuming uniform priors is equivalent to an application
	of the method of moments. Then, using synthetic data, we show that the Bayesian estimate scales poorly with number of nodes in the network
	as compared to an iterative Lyapunov optimization. In particular when the network size is in the order of that
	used for whole-brain studies (about 100 nodes) the Bayesian method needs about eight times more time samples than Lyapunov method
	in order to achieve similar estimation accuracy. 
	We also show that the higher estimation accuracy of Lyapunov method is reflected in a much better classification of individuals based on the
	estimated connectivity from a real dataset of BOLD fMRI.
	Finally we show that the poor accuracy of Bayesian method is due to numerical errors, when the imaginary part of the connectivity estimate gets
	large compared to its real part.
\end{abstract}

\section*{Introduction}

Estimation of connectivity from time series data is a common goal for many disciplines, such as economics, system biology or neuroscience.
In neuroscience there has been increasing interest over the last years for connectivity measures at the whole-brain scale as defined by recording
techniques like
fMRI or EEG. Examples of connectivity measures include statistical dependence measures such as Pearson's correlation or mutual information and
network models such as vector autoregressive models, dynamic causal model \citep{friston_functional_2011} or multivariate Ornstein-Uhlenbeck
\citep{gilson_estimation_2016}. 

The estimated connectivity can then be used to study individual traits \citep{finn_functional_2015,pallares_subject-_2017}, cognitive states
\citep{gonzalez-castillo_tracking_2015,pallares_subject-_2017} or clinical conditions \citep{rahim_joint_2017}. In particular connectivity
measures open new perspectives as biomarkers for neuropsychiatric diseases \citep{arbabshirani_single_2017,greicius_resting-state_2008,meng_predicting_2017-1}.

Estimating brain connectivity is in general challenging due to the scarcity of time samples (in the order of hundreds for fMRI) and the
large dimensionality of the measurements (typical brain parcellation about 100 regions or more). 
Bayesian estimation could bring some advantages such as the ability to naturally perform connectivity detection, by making statistical tests exploiting
the posterior probability distribution of the connectivity. In addition the
inclusion of a prior over model parameters can act as a regularization term and in turn improve estimation accuracy when only few samples are available.
Finally the Bayesian framework naturally allows to extend models to include modelling of population variability through hierarchical models and different observational models 
\citep{linderman_bayesian_2016}.

Here we study a Bayesian estimation method for the multivariate Ornstein-Uhlenbeck model (mOU) and compare it with other available estimation methods: the moments method and a
Lyapunov optimization \citep{gilson_estimation_2016}. 

\section{Model definition}

In this work we consider a multivariate Ornstein-Uhlenbeck (mOU) process. 
The process instantiates a network model where the dynamics of each node is governed by an OU equation with additional input from all connected nodes: 

\begin{equation}
	\label{mou}
\frac{dx_i}{dt} = -\frac{x_i}{\tau_x} + \sum_{j\ne i}^M C_{ij} x_j + \sigma_i \frac{dB_i}{dt}
\end{equation}

where $x_i$ is the activity of the $i$th node, $\tau_x$ is the time constant of the node, $C$ is the connectivity matrix where each entry $i,j$
represent the link strength from node $j$ to node $i$ (the diagonal of the $C$ is set to 0 to avoid self connections), $\sigma_i$ is the noise variance of each node that scales the Gaussian noise $dB_i/dt$. All 
$\sigma_i^2$ can be collected in the diagonal of noise covariance matrix $\Sigma$. Here we assume uncorrelated noise, hence $\Sigma$ is diagonal,
but the model can be easyly generalized to correlated noise.
We also assume the time constant $\tau_x$ to be omogeneous between nodes. Here we are not consider any additional node specific term (usually
called drift). However we note that a drift term could be introduced into the equation and its value estimated for decoding purposes,
e.g. in scenarios where external stimuli are relevant.

\section{Model estimation}

In this section, following \citet{gilson_estimation_2016} and \citet{singh_fast_2017} we first show how to calculate model covariance knowing the parameters then we outline the basic steps to estimate
mOU parameters using three different procedures: moments method, Lyapunov optimization and Bayesian posterior mean.

\subsection{Forward step}
The Jacobian of the system is: $ J_{ij} = -\frac{\delta_{ij}}{\tau_x} + C_{ij}$, where $\delta_{ij}$ is Kronecker's delta.
Deriving the covariance of the system we obtain the Lyapunov equation:
\begin{equation}
	JQ^0 + Q^0J^T + \Sigma = 0
	\label{lyap}
\end{equation}
where $\Sigma$ is the diagonal matrix
$\Sigma_{ii} = \sigma_i^2$.
From this equation the covariance can be calculated using the Bartels-Stewart algorithm \citep{bartels_solution_1972}. 

Similarly the derivation of the time-lagged covariance for a given lag $\tau$ yealds:
\begin{equation}
\label{qtau}
 Q^{\tau} = Q^0 expm(J^T\tau)
\end{equation}
where $expm(A)$ is the the matrix exponential of $A$.

Hence knowing $Q^0$, $Q^{\tau}$ can be calculated.

These equations allow to compute the farward step: calculate (time-lagged) covariance matrix knowing the parameters of the system $\theta: \{C, \Sigma, \tau_x\}$.

\subsection{Moments method}

As shown by \citet{gilson_estimation_2016}, the moments method can be applied in order to estimate the parameters $\theta$ (called direct estimation in \citet{gilson_estimation_2016}).
We can substitute the theoretical covariances in eq. \eqref{qtau} for their empirical counterparts $\hat{Q}^0$, $\hat{Q}^{\tau}$ and 
solve for $J$ to get:
\begin{equation}
\label{}
 \hat{J} = \frac{1}{\tau} \Bigl[logm(\hat{Q}^{\tau}(\hat{Q}^0)^{-1})\Bigr]^{T}  
\end{equation}
where $logm(A)$ is the matrix logarithm of $A$.

Then with the estimate $\hat{J}$, an estimate of $\Sigma$ can then be calculated from Lyapunov equation \eqref{lyap}:
\begin{equation}
\label{}
 \hat{\Sigma} = -\hat{J}\hat{Q}^0 - \hat{Q}^0 \hat{J}^T 
\end{equation}

\subsection{Lyapunov optimization}
\citet{gilson_estimation_2016} developed a Lyapunov optimization for the parameters of mOU. This optimization procedure minimizes the Lyapunov
function:
\begin{equation}
\label{}
V(C) = \sum_{m,n}(Q_{mn}^0-\hat{Q}_{mn}^0)^2 + \sum_{m,n}(Q_{mn}^{\tau}-\hat{Q}_{mn}^{\tau})^2 
\end{equation}

By differentiating this function the parameters update can be obtained as:
\begin{equation}
\label{}
 \Delta J = \frac{1}{\tau} \Bigl[(Q^0)^{-1} (\Delta Q^0 + \Delta Q^{\tau} expm(-J^T \tau)) \Bigr]^T
\end{equation}

\begin{equation}
\label{}
\Delta \Sigma = -\hat{J}\Delta Q^0 - \Delta Q^0 \hat{J}^T
\end{equation}

The reader is referenced to \citet{gilson_estimation_2016} for further details.

\subsection{Bayesian approach}
Here we follow \citet{singh_fast_2017} in the definition of a Bayesian estimate.

The probability of the state of the system at time $t'$ given its state at time $t$ is given by:
\begin{equation}
\label{}
 x(t'|t) \sim \mathcal{N}\Bigl(expm(J\Delta t)x(t), \Xi \Bigr)
\end{equation}

where $\Xi = Q^0 - expm(J\Delta t) Q^0 expm(J \Delta t)^T$ and $\Delta t = t'-t$.

The stationary distribution is: $x \sim \mathcal{N}(0, Q^0)$.
We recall that system's covariance matrix $Q^0$ is related to $\Sigma$ by Lyapunov equation: $ JQ^0 + Q^0J^T + \Sigma = 0 $.

Given a dataset where x is sampled at regular intervals $x^1, x^2,\dots, x^N$, and $X$ is the $M$ times $N$ matrix collecting all the $N$ observations for each of the $M$ nodes, the likelihood function is given by:
\begin{equation}
\label{}
p(X|\theta) = \prod_n^N p(x^{n+1} | x^n, \theta)p(x^1|\theta)
\end{equation}

Then the posterior distribution of parameters is given by Bayes theorem:
\begin{equation}
\label{}
 p(\theta|X)=\frac{p(X|\theta)p(\theta)}{p(X)}
\end{equation}

Assuming a uniform prior over the paramters and substituting the explicit pdfs the log posterior is:
\begin{equation}
\label{logpost}
 ln\, p(\theta|X) = -\frac{1}{2} \sum_n^N \Delta_n^T \Xi \Delta_n -\frac{1}{2} (x^1)^T (Q^0)^{-1} x^1 + \frac{N-1}{2} ln \frac{1}{(2\pi)^M |\Xi|} + \frac{1}{2} ln \frac{1}{(2\pi)^M |Q^0|} 
\end{equation}

where $\Delta_n = x^{n+1}-\Lambda x^n$ and $\Lambda=expm(J \Delta t)$.

The first term of \eqref{logpost} can be expanded to show that the log posterior is normal in $\Lambda$.
Finally, defining $T^0 = \sum^N x^n(x^n)^T $ and $T^1 = \sum^N x^{n+1}(x^n)^T $, the posterior mean for $\Lambda$ can be written as:
$\hat{\Lambda} = T^1 (T^0)^{-1}$ ($\Xi$ can also be estimated with the same method but we don't pursue it here since it is not related to connectivity estimation).

It follows than that:
\begin{equation}
\label{}
 \hat{J} = \frac{logm(\hat{\Lambda})}{\Delta t} 
\end{equation}

Sigma can then be estimated again from the Lyapunov equation:
\begin{equation}
\label{}
 \hat{\Sigma} = -\hat{J}\hat{Q}^0 - \hat{Q}^0 \hat{J}^T 
\end{equation}

By noting that $T^0$ and $T^1$ are proportional to the empirical covariance matrix and the transposed empirical lagged covariance matrix it can be easily seen
that the Bayesian and the moments methods yield the same solution.

\section{Estimation accuracy for large scale systems}

Here we study the accuracy of the estimation methods presented above in the context of systems where the number of variables is large.
We first study the estimation accuracy of synthetic data, where the ground truth is known and then move to the estimation with empirical data.

\subsection{Synthetic network with random connectivity}

Synthetic data are generated by simulating a mOU process. To this aim we used Euler integration with time step of 0.05, which is small compared to
the time constant $\tau_x$. The resulting simulated time series are then downsampled to 1 second (to have similar
time resolution as typical fMRI recordings). For the simulations we fixed $\tau_x=1$ second and $\Sigma$ to a diagonal matrix $diag(0.5 + 0.5*\sigma)$, where $\sigma$ is sampled from a uniform distribution
between 0 and 1.

Connectivity matrix $C$ was constructed as the Hadamard (element-wise) product of a binary adjacency matrix and a log-normal weight matrix.

\begin{equation}
\label{}
 C' = A \odot W 
\end{equation}

where $ A\sim Bern(p)$ and $ ln W \sim \mathcal{N}(0,1)$

Thus the underlying network is Erd\H{o}s-R\'enyi while the strength of each link is sampled from a log-normal distribution.
While this corresponds to assuming a very simple model we show below that our results hold also for more structured networks.

When varying the number of nodes C' gets normalized in order to avoid explosion of activity.
\begin{equation}
\label{}
 C = \frac{C' M}{\sum_{i,j} C'} 
\end{equation}

In general, given a fixed amount of samples, the more parsimonious (i.e. less parameters) a model the higher the precision in the estimate of its 
parameters. The model under consideration has $M(M+1)$ parameters (including $\Sigma$ and $\tau_x$), where $M$ is the number of nodes in the network.

\begin{figure}[htpb]
	\centering
	\includegraphics[width=0.99\columnwidth]{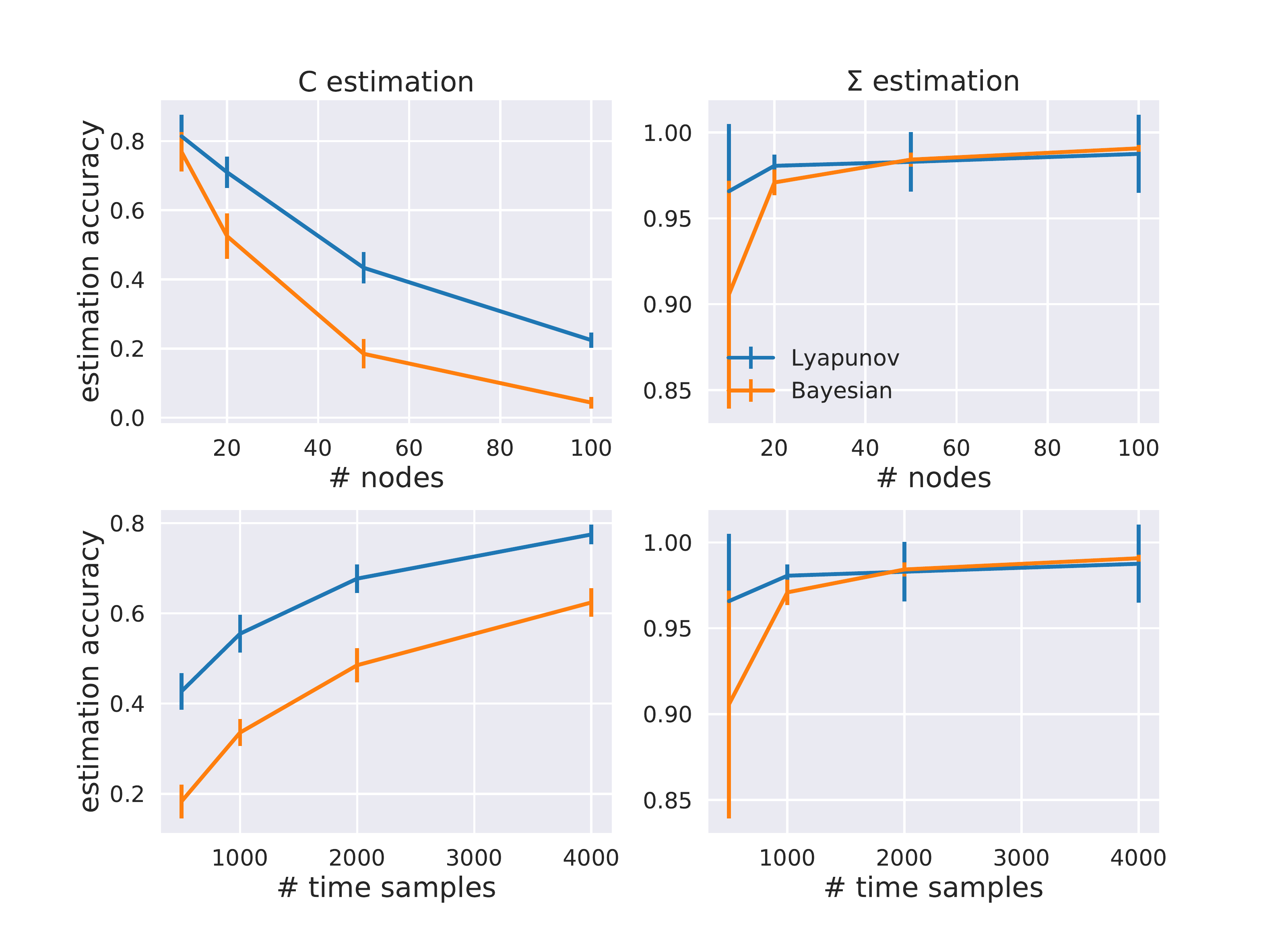}
	\caption{Estimation accuracy with a random generative connectivity matrix. Accuracy for $C$ (left column) and $\Sigma$ (rigth column), varying the number of nodes $M$ in the network (upper row, withfixed $N=500$) or the number of time samples $N$ (bottom row, with fixed $M=50$). The estimation is performed with Lyapunov (blu curves) or Bayesian methods (orange curves).}
	\label{fig:estim_acc_M}
\end{figure}

We show here how the size of the network $M$ and the number of time samples $N$ influences the accuracy of the estimation.
For a given size $M$ we draw a connectivity matrix and simulate the model for
500 seconds. Then we estimate the connectivity from the simulated time
series using Lyapunov and Bayesian methods. Finally, we calculate Pearson's
correlation coefficient between the true connectivity (used to generate the
time series) and the estimated connectivity, as a measure of the accuracy of
estimation. We repeat these steps 100 times for each value of $M$.
As expected, the largest effect is on the estimation of $C$ as illustrated in fig. \ref{fig:estim_acc_M}. The accuracy for the estimation of $\Sigma$ presents a small decrease for Bayesian method and no notable modulation for 
Lyapunov method. The accuracy in the estimate of $C$ decreases as a function of network size for both estimation methods. However Lyapunov estimation shows a slower decrease
of accuracy and allows more reliable estimates.

To complement this analysis we also show the effect of the amount of time samples for a given network size.
For a each number of time samples $N$ we draw a connectivity matrix and
simulate the model with 50 nodes. Then we estimate the connectivity using Lyapunov and Bayesian methods and evaluate estimation accuracy as above.
As can be observed in fig. \ref{fig:estim_acc_M} Lyapunov estimation needs about four times fewer samples to achieve similar accuracy as Bayesian method for the estimation of $C$. The estimation of $\Sigma$ shows a
small increase in accuracy for both estimation methods (notice that the accuracy is already very high for $N=500$).

\subsection{Synthetic network with connectivity estimated from empirical data}

\begin{figure}[htpb]
	\centering
	\includegraphics[width=0.9\columnwidth]{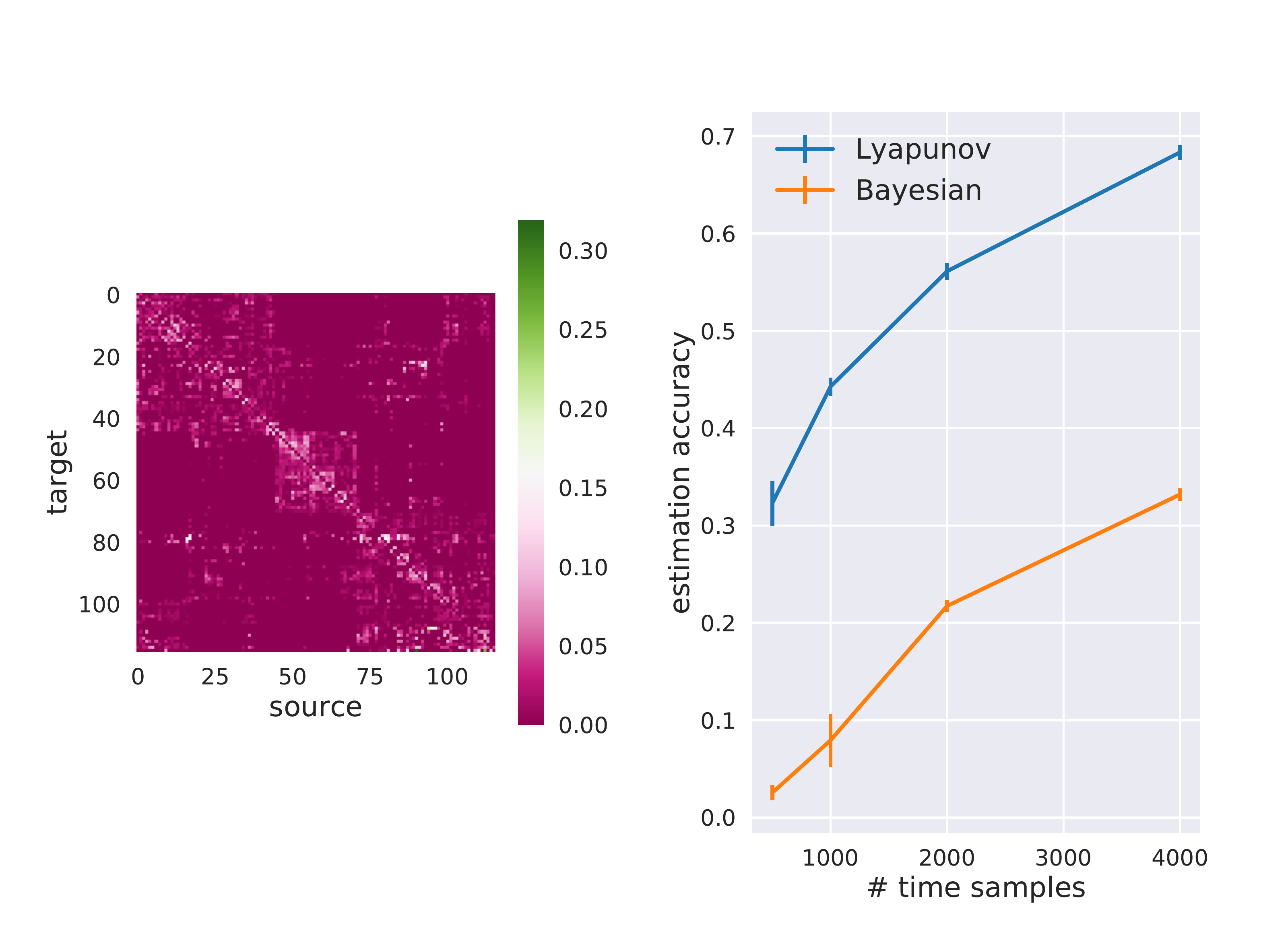}
	\caption{Left: Connectivity matrix estimated from BOLD fMRI data. Right: Estimation accuracy of $C$ for time series simulated using
	the connectivity matrix in left panel.}
	\label{fig:estim_acc_T_emp}
\end{figure}

Typical estimate of brain structural and functional connectivity show significantly more structure than Erd\H{o}s-R\'eniy networks.
Here we show that the above results also hold for a real brain topology obtained from anatomical contraints. 
In order to obtain a generative model
similar to experimental data we estimate the connectivity matrix from the BOLD time series of an fMRI recording session, using Lyapunov optimization
and a structural connectivity template (from diffusion tensor imaging) as a mask for the estimation. The BOLD time series used was taken 
from \citet{zuo_open_2014} session 1 of subject with id 25427 preprocess with AAL parcellation ($M=116$ nodes) 
as described in \citet{pallares_subject-_2017}. The structural connectivity corresponds to a generic template (obtained from other subjects) for
the AAL parcellation \citep{tzourio-mazoyer_automated_2002}, which enforced a non-random topology (see left panel in fig. \ref{fig:estim_acc_T_emp}).
We then use this connectivity matrix to generate simulated time series using the mOU as above.

In fig. \ref{fig:estim_acc_T_emp} right we show the influence of the amount of time samples on the estimation accuracy.
For a given number of time samples $N$ we simulate the model using the connectivity matrix estimated from empirical data 
(fig. \ref{fig:estim_acc_T_emp}). Then we estimate the connectivity from the simulated time series using Lyapunov and Bayesian methods and evaluate
estimation accuracy as above.

It can be observed that the accuracy increases as a function of the number of time samples but Lyapunov method has always a higher accuracy and also 
a faster increase compared to Bayesian method. We note that, as shown in fig. \ref{fig:estim_acc_M} Bayesian method suffers more from high dimensionality of the
system; as a result here ($M=116$) Bayesian needs 8 times more time sample to achieve a similar accuracy as Lyapunov method.

\section{Higher estimation accuracy is reflected in better subjects identification}

There is an increasing interest in measure of brain connectivity for identifying subjects \citep{finn_functional_2015,pallares_subject-_2017},
cognitive states \citep{gonzalez-castillo_tracking_2015,pallares_subject-_2017} or clinical conditions \citep{rahim_joint_2017}.
From the previous results we expect the Lyapunov method to extract more informative parameters value than Bayesian, which should be in turn advantageous for classification. 

In the following we show how the Lyapunov and Bayesian estimates of connectivity
perform in a classification task of subjects identity. For this we used a
dataset where 30 subject underwent 10 fMRI resting state scanning sessions
\citet{zuo_open_2014}.
For each subject and session we estimate the connectivity with either the
Bayesian or Lypunov method. Here we use the generic structural connectivity template \citep{tzourio-mazoyer_automated_2002} to mask the estimated value of $C$ where there is no connection in the structural 
connectivity. We then classify the identity of the subjects
using the values of estimated connectivity.
To this aim our data for classification is composed of 300 points (one for each subject and session) in 4056 dimensions (all non-zero values in the structural connectivity matrix).
We randomly split the data using 80\% as training set and the rest for testing. We train a multinomial logistic regression classifier on the training set and evaluat its performance on the test set.
This procedure is repeated 100 times. For a more complete study of subject and behavior classification see \citet{pallares_subject-_2017}.

\begin{figure}[htpb]
	\centering
	\includegraphics[width=0.8\columnwidth]{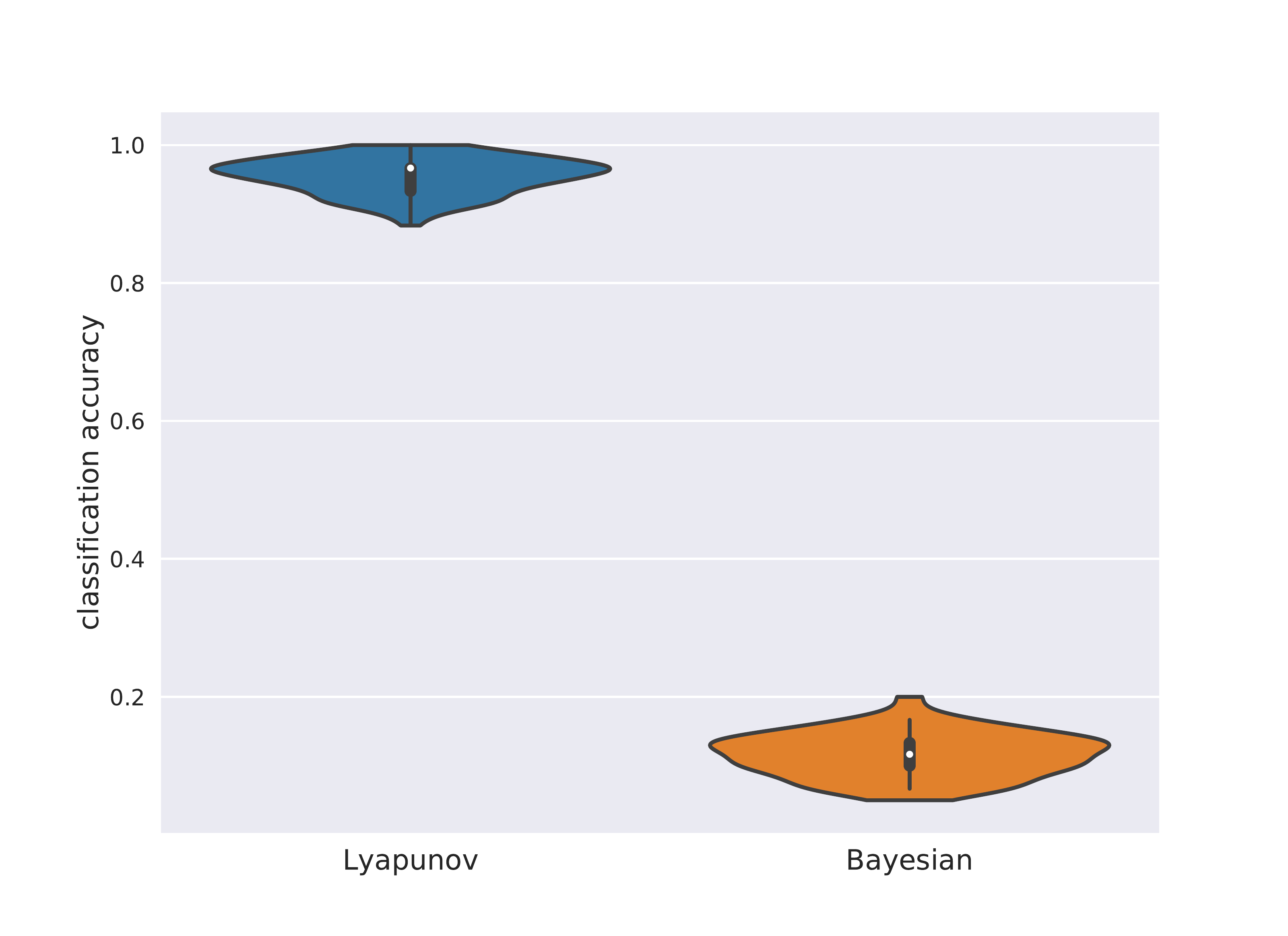}
	\caption{Distribution of classification accuracy for 100 test-sets comprising 20\% of randomly chosen samples for the classification of subjects identity based on the value of links in the connectivity matrix estimated either with Lyapunov or with Bayesian method.}
	\label{fig:classif}
\end{figure}

Fig. \ref{fig:classif} shows the distribution of classifier accuracy as violin plots. As can be observed the higher reliability of lyapunov method is reflected in higher classification accuracy.

\section{Responsibility of estimation error for Bayesian method}

\begin{figure}[htpb]
	\centering
	\includegraphics[width=0.9\columnwidth]{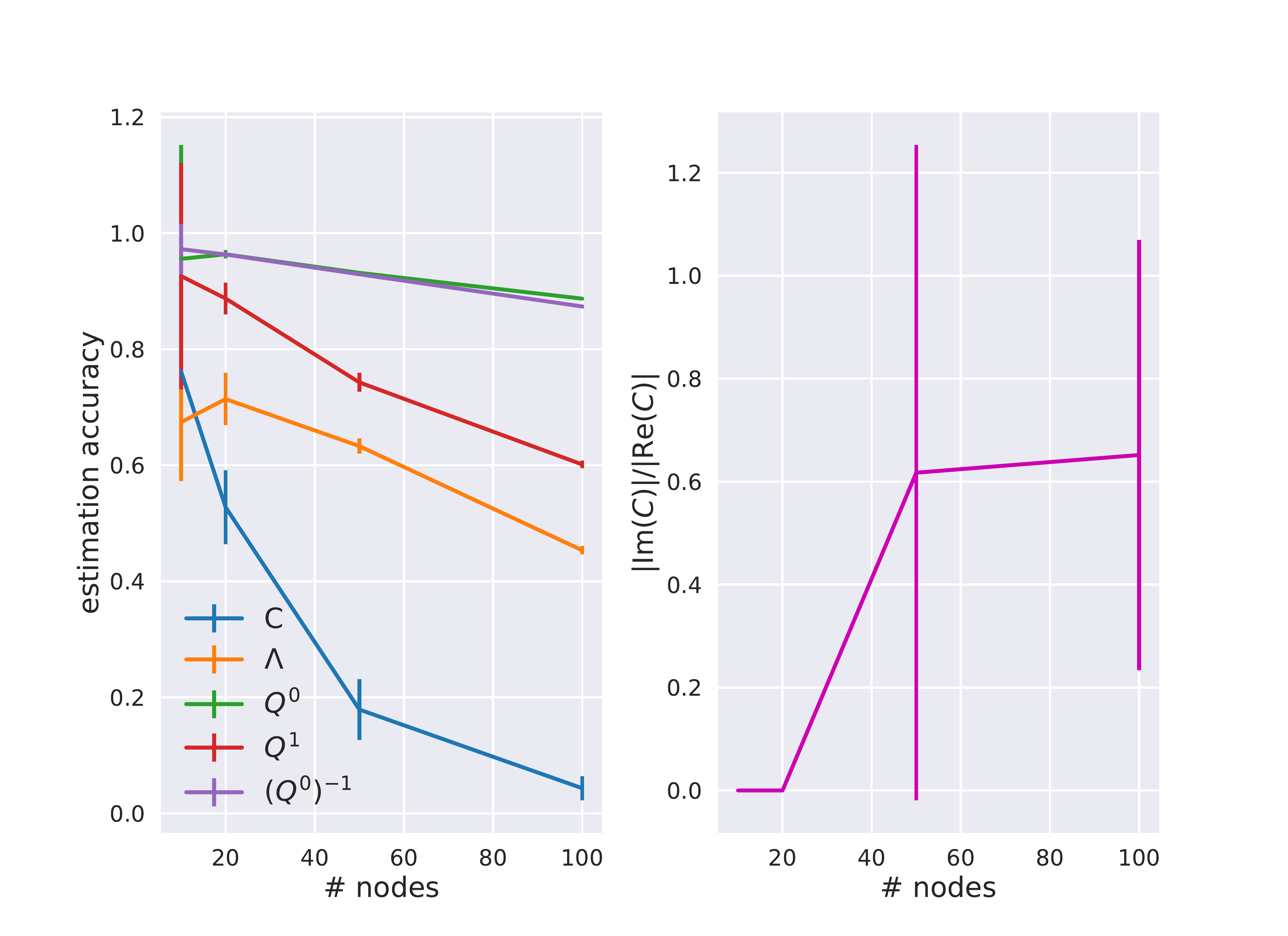}
	\caption{Left: Estimation accuracy for intermediate steps in the calculation of Bayesian estimate as a function of the number of nodes in the network. The estimate $\hat{C}$ is compared with the generative connectivity $C$, while all other estimates are compared to their theoretical equivalents calculated from the parameters. Right: Ratio of the norms of the imaginary and real part of connectivity $C$ as a function of the number of nodes in the network. Errorbars show standard deviation over 100 simulations.}
	\label{fig:why_fail}
\end{figure}

The Bayesian method is based on the matrix logarithm of $\hat{\Lambda} = (\hat{Q}^0)^{-1} \hat{Q}^{\tau}$, where precision matrix $(\hat{Q}^0)^{-1}$ and lagged-covariance $\hat{Q}^{\tau}$
are both estimated from the time series.
The large estimation error shown above is presumably due to numerical errors. In order to understand which part of this estimation has the largest responsibility in the estimation error, in fig. \ref{fig:why_fail} 
we show the similarity (measured by Pearson's correlation) between the estimated and theoretical counterparts
of the different steps in the estimation of connectivity. As above we simulate the mOU for varying network size and compare the estimate
$\hat{C}$ with the generative connectivity $C$ and the estimates $(\hat{Q}^0)^{-1}$, $\hat{Q}^{\tau}$, $\hat{\Lambda}$ and $logm(\hat{\Lambda})$ with their theoretical counterparts calculated using the
parameters of the generative model.

It can be observed in left panel that the estimation of lagged-covariance suffers much more than covariance and precision from the increase of network size. This gets also reflected in the similarity of $\hat{\Lambda}$ with its theoretical counterpart.
However the largest responsibility for estimation error comes from the matrix logarithm (blue line in fig. \ref{fig:why_fail} left).
Indeed when number of nodes increases the imaginary part of $\hat{C}$ becomes larger compared to its real part as shown in 
fig. \ref{fig:why_fail} right.

\section{Discussion}

Here we show that Bayesian estimate for mOU model is equivalent to an estimation based on moments method. We also show that Bayesian method is
more sensible to network size and number of time samples than Lyapunov method.
A number of improvement could be done for the Bayesian estimation (and will form the basis for future work).
First, a different prior could be used for the Bayesian estimate. Here we used a uniform prior, so a natural extension could be a Gaussian prior
that would act as a regularization term, thereby probably increasing the accuracy for small number of time samples. We note that the Lyapunov estimation
used here could be easily extended to also include a regularization term.

Another option whould be the use of an empirical prior, for example derived from structural connectivity or covariance matrix of BOLD time series, possibly in conjunction with an expectation-maximization
algorithm. Indeed, while we used the structural connectivity matrix to mask estimated values, these are not excluded from the estimation procedure. On the contrary, the iterative Lyapunov optimization clips to
zero those links that are zero in the structural connectivity, thereby allowing other links to vary more. Hence the structural connectivity acts like a sort of empirical prior in the Lyapunov estimation 
procedure.

Finally another natural extension of the Bayesian estimation would be to perform link detection to filter out links that are estimated non-zero merely as an effect of noise.
This would produce a sparse estimate for the connectivity and presumably increase the estimation accuracy.

\subsubsection*{Acknowledgments}

\bibliographystyle{apalike}
\bibliography{../BIB/myzoterobib.bib} %

\begin{thebibliography}{}

\bibitem[Arbabshirani et~al., 2017]{arbabshirani_single_2017}
Arbabshirani, M.~R., Plis, S., Sui, J., and Calhoun, V.~D. (2017).
\newblock Single subject prediction of brain disorders in neuroimaging:
  {Promises} and pitfalls.
\newblock {\em NeuroImage}, 145:137--165.

\bibitem[Bartels and Stewart, 1972]{bartels_solution_1972}
Bartels, R.~H. and Stewart, G.~W. (1972).
\newblock Solution of the {Matrix} {Equation} {AX} + {XB} = {C} [{F}4].
\newblock {\em Commun. ACM}, 15(9):820--826.

\bibitem[Finn et~al., 2015]{finn_functional_2015}
Finn, E.~S., Shen, X., Scheinost, D., Rosenberg, M.~D., Huang, J., Chun, M.~M.,
  Papademetris, X., and Constable, R.~T. (2015).
\newblock Functional connectome fingerprinting: identifying individuals using
  patterns of brain connectivity.
\newblock {\em Nature Neuroscience}, 18(11):1664--1671.

\bibitem[Friston, 2011]{friston_functional_2011}
Friston, K.~J. (2011).
\newblock Functional and {Effective} {Connectivity}: {A} {Review}.
\newblock {\em Brain Connectivity}, 1(1):13--36.

\bibitem[Gilson et~al., 2016]{gilson_estimation_2016}
Gilson, M., Moreno-Bote, R., Ponce-Alvarez, A., Ritter, P., and Deco, G.
  (2016).
\newblock Estimation of {Directed} {Effective} {Connectivity} from {fMRI}
  {Functional} {Connectivity} {Hints} at {Asymmetries} of {Cortical}
  {Connectome}.
\newblock {\em PLOS Computational Biology}, 12(3):e1004762.

\bibitem[Gonzalez-Castillo et~al., 2015]{gonzalez-castillo_tracking_2015}
Gonzalez-Castillo, J., Hoy, C.~W., Handwerker, D.~A., Robinson, M.~E.,
  Buchanan, L.~C., Saad, Z.~S., and Bandettini, P.~A. (2015).
\newblock Tracking ongoing cognition in individuals using brief, whole-brain
  functional connectivity patterns.
\newblock {\em Proceedings of the National Academy of Sciences},
  112(28):8762--8767.

\bibitem[Greicius, 2008]{greicius_resting-state_2008}
Greicius, M. (2008).
\newblock Resting-state functional connectivity in neuropsychiatric disorders.
\newblock {\em Current Opinion in Neurology}, 21(4):424.

\bibitem[Linderman et~al., 2016]{linderman_bayesian_2016}
Linderman, S., Adams, R.~P., and Pillow, J.~W. (2016).
\newblock Bayesian latent structure discovery from multi-neuron recordings.
\newblock In Lee, D.~D., Sugiyama, M., Luxburg, U.~V., Guyon, I., and Garnett,
  R., editors, {\em Advances in {Neural} {Information} {Processing} {Systems}
  29}, pages 2002--2010. Curran Associates, Inc.

\bibitem[Meng et~al., 2017]{meng_predicting_2017-1}
Meng, X., Jiang, R., Lin, D., Bustillo, J., Jones, T., Chen, J., Yu, Q., Du,
  Y., Zhang, Y., Jiang, T., Sui, J., and Calhoun, V.~D. (2017).
\newblock Predicting individualized clinical measures by a generalized
  prediction framework and multimodal fusion of {MRI} data.
\newblock {\em NeuroImage}, 145:218--229.

\bibitem[Pallares et~al., 2017]{pallares_subject-_2017}
Pallares, V.~., Insabato, A.~., Sanjuan, A., Kuehn, S., Mantini, D., Deco, G.,
  and Gilson, M. (2017).
\newblock Subject- and behavior-specific signatures extracted from {fMRI} data
  using whole-brain effective connectivity.
\newblock {\em bioRxiv}, page 201624.

\bibitem[Rahim et~al., 2017]{rahim_joint_2017}
Rahim, M., Thirion, B., Bzdok, D., Buvat, I., and Varoquaux, G. (2017).
\newblock Joint prediction of multiple scores captures better individual traits
  from brain images.
\newblock {\em NeuroImage}, 158:145--154.

\bibitem[Singh et~al., 2017]{singh_fast_2017}
Singh, R., Ghosh, D., and Adhikari, R. (2017).
\newblock Fast {Bayesian} inference of the multivariate {Ornstein}-{Uhlenbeck}
  process.
\newblock {\em arXiv:1706.04961 [cond-mat, physics:physics]}.
\newblock arXiv: 1706.04961.

\bibitem[Tzourio-Mazoyer et~al., 2002]{tzourio-mazoyer_automated_2002}
Tzourio-Mazoyer, N., Landeau, B., Papathanassiou, D., Crivello, F., Etard, O.,
  Delcroix, N., Mazoyer, B., and Joliot, M. (2002).
\newblock Automated {Anatomical} {Labeling} of {Activations} in {SPM} {Using} a
  {Macroscopic} {Anatomical} {Parcellation} of the {MNI} {MRI}
  {Single}-{Subject} {Brain}.
\newblock {\em NeuroImage}, 15(1):273--289.

\bibitem[Zuo et~al., 2014]{zuo_open_2014}
Zuo, X.-N., Anderson, J.~S., Bellec, P., Birn, R.~M., Biswal, B.~B., Blautzik,
  J., Breitner, J. C.~S., Buckner, R.~L., Calhoun, V.~D., Castellanos, F.~X.,
  Chen, A., Chen, B., Chen, J., Chen, X., Colcombe, S.~J., Courtney, W.,
  Craddock, R.~C., Di~Martino, A., Dong, H.-M., Fu, X., Gong, Q., Gorgolewski,
  K.~J., Han, Y., He, Y., He, Y., Ho, E., Holmes, A., Hou, X.-H., Huckins, J.,
  Jiang, T., Jiang, Y., Kelley, W., Kelly, C., King, M., LaConte, S.~M.,
  Lainhart, J.~E., Lei, X., Li, H.-J., Li, K., Li, K., Lin, Q., Liu, D., Liu,
  J., Liu, X., Liu, Y., Lu, G., Lu, J., Luna, B., Luo, J., Lurie, D., Mao, Y.,
  Margulies, D.~S., Mayer, A.~R., Meindl, T., Meyerand, M.~E., Nan, W.,
  Nielsen, J.~A., O’Connor, D., Paulsen, D., Prabhakaran, V., Qi, Z., Qiu,
  J., Shao, C., Shehzad, Z., Tang, W., Villringer, A., Wang, H., Wang, K., Wei,
  D., Wei, G.-X., Weng, X.-C., Wu, X., Xu, T., Yang, N., Yang, Z., Zang, Y.-F.,
  Zhang, L., Zhang, Q., Zhang, Z., Zhang, Z., Zhao, K., Zhen, Z., Zhou, Y.,
  Zhu, X.-T., and Milham, M.~P. (2014).
\newblock An open science resource for establishing reliability and
  reproducibility in functional connectomics.
\newblock {\em Scientific Data}, 1:140049.

\end{thebibliography}

\end{document}